\documentclass[final]{cvpr}
\usepackage{url}
\usepackage{times}
\usepackage{epsfig}
\usepackage{graphicx}
\usepackage{amsmath}
\usepackage{amssymb}
\usepackage{color}
\usepackage{grffile}
\usepackage{tablefootnote}
\usepackage{cases}
\usepackage{sidecap}
\usepackage{wrapfig}
\usepackage{float}
\usepackage{leftidx}
\usepackage{pifont}% http://ctan.org/pkg/pifont

\usepackage{pifont}
\usepackage{algorithmic,algorithm}
\usepackage{bm}
\usepackage[square, comma, sort&compress, numbers]{natbib} % For continuous reference numbers
\usepackage[english]{babel} % American English
\usepackage[shortlabels]{enumitem}
\usepackage[pagebackref=true,breaklinks=true,letterpaper=true,colorlinks,bookmarks=false]{hyperref}
\usepackage{booktabs}
\usepackage{colortbl}
\usepackage{subcaption}
\usepackage{multirow}
\usepackage{makecell}
\usepackage{gensymb}
\usepackage{pifont}
\usepackage{dashrule}
\usepackage[export]{adjustbox}
\usepackage{graphbox,graphicx}
\usepackage{threeparttable}
\usepackage{arydshln}

\usepackage{hhline}
\usepackage{siunitx}
\RequirePackage{enumitem}
\setlist[itemize]{nosep}

\newcommand{\Tref}[1]{Table~\ref{#1}}
\newcommand{\eref}[1]{Eq.~\eqref{#1}}

\newcommand{\fref}[1]{Fig.~\ref{#1}}

\newcommand{\sref}[1]{Sec.~\ref{#1}}

\usepackage{xcolor}
\newcounter{todos}
\AtEndDocument{\ifnum\value{todos}>0 \PackageWarning{TODOS}{There are \arabic{todos} todos left in this paper! Fix them before submitting the paper!} \fi}

\newcommand{\V}[1]{\ensuremath{\mathbf{#1}}}

\usepackage[pagebackref=true,breaklinks=true,colorlinks,bookmarks=false]{hyperref}

\renewcommand{\paragraph}[1]{\vspace{1.0em}\noindent \textbf{#1 \hspace{0.2em}}}

\newcommand{\lightNet}{ELIE-Net\xspace}
\newcommand{\normalModule}{SNE module\xspace}

\def\cvprPaperID{3927} % *** Enter the CVPR Paper ID here
\def\confYear{CVPR 2022}

\definecolor{Gray}{gray}{0.9}
\newcommand\blfootnote[1]{%
	\begingroup
	\renewcommand\thefootnote{}\footnote{#1}%
	\addtocounter{footnote}{-1}%
	\endgroup
}

\makeatletter
\newcommand*{\rom}[1]{\expandafter\@slowromancap\romannumeral #1@}
\makeatother
 
\begin{document}
% 	\pagenumbering{gobble}
	%%%%%%%%% TITLE
	\title{NeuralMPS: Non-Lambertian Multispectral Photometric Stereo via \\
	Spectral Reflectance Decomposition}
	\author{
            Jipeng Lv$^{1,*}$ \quad
		Heng Guo$^{2,*}$ \quad
		Guanying Chen$^3$ \quad
		Jinxiu Liang$^1$ \quad
		Boxin Shi$^1$ \\ 
		$^1$Peking University
		$^2$Osaka University 
	    $^3$The Chinese University of Hong Kong, Shenzhen
	}
	
	\maketitle

	%%%%%%%%% ABSTRACT
	\begin{abstract}
		%\vspace{-8pt}
		% what's the problem
		Multispectral photometric stereo~(MPS) aims at recovering the surface normal of a scene from a single-shot multispectral image captured under multispectral illuminations. Existing MPS methods adopt the Lambertian reflectance model to make the problem tractable, but it greatly limits their application to real-world surfaces. In this paper, we propose a deep neural network named NeuralMPS to solve the MPS problem under general non-Lambertian spectral reflectances. Specifically, we present a spectral reflectance decomposition~(SRD) model to disentangle the spectral reflectance into geometric components and spectral components. With this decomposition, we show that the MPS problem for surfaces with a uniform material is equivalent to the conventional photometric stereo~(CPS) with unknown light intensities. In this way, NeuralMPS reduces the difficulty of the non-Lambertian MPS problem by leveraging the well-studied non-Lambertian CPS methods. Experiments on both synthetic and real-world scenes demonstrate the effectiveness of our method.
	\end{abstract}
        \blfootnote{*These authors contributed equally to this work.}

	\section{Introduction}
	\label{sec:intro}
	% step 1: what is the task, background
	\newcommand\imgsize{0.2}
\begin{figure}\centering
	\includegraphics[width=\linewidth]{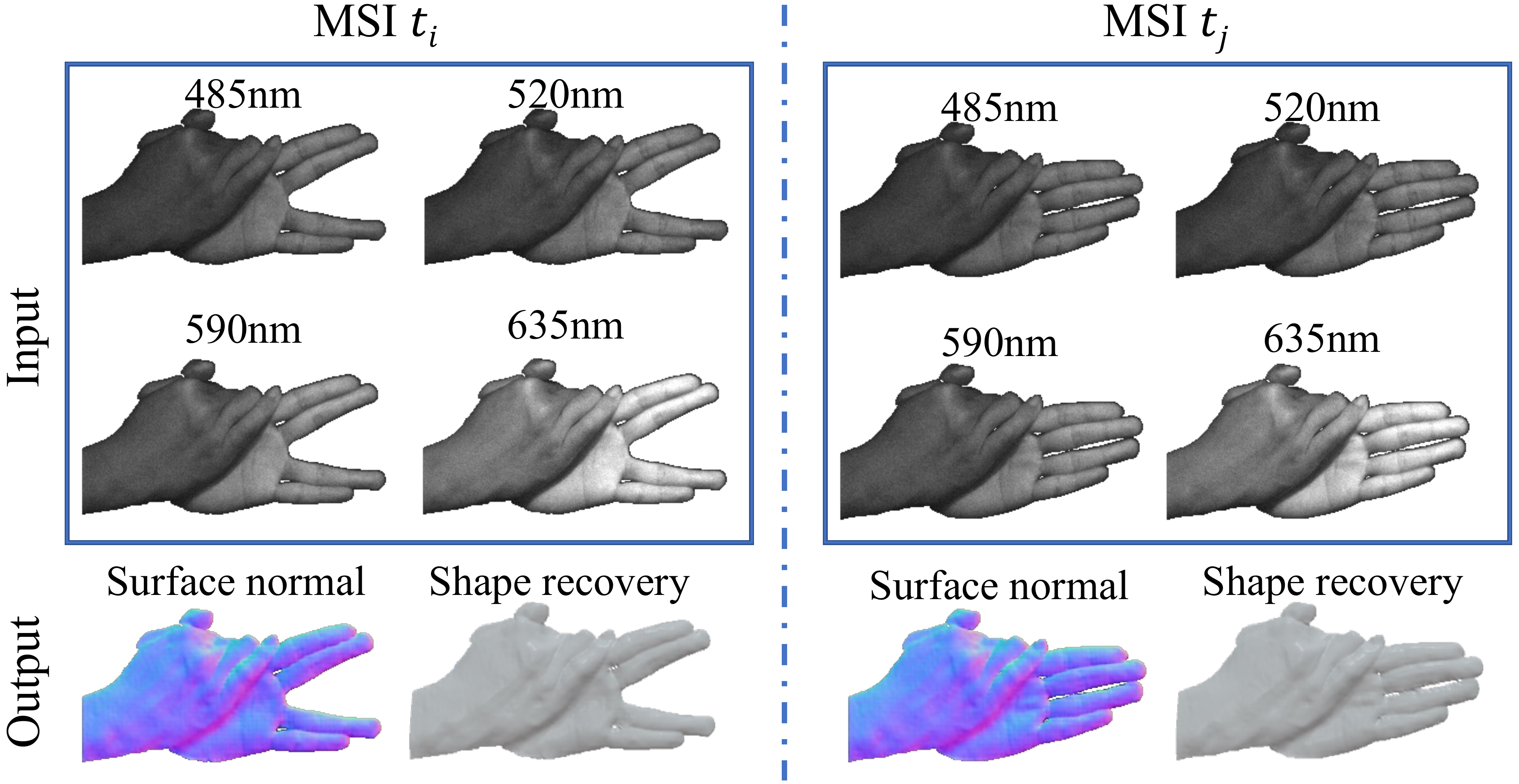}
	\caption{Multispectral image~(MSI) stacks image observations under varying lighting directions at different spectral bands. Our multispectral photometric stereo (MPS) method takes MSI at different timestamps\protect\footnotemark and recovers the dynamic 3D shape~(surface normal). }
	\label{fig:teaser}
\end{figure}

	Photometric stereo methods, originally proposed by Woodham~\cite{woodham1980ps} and Silver~\cite{silver1980determining}, recover detailed geometry of three-dimensional (3D) surfaces from images captured from a fixed camera under varying lighting directions, which are commonly obtained at different timestamps.
	Since conventional photometric stereo~(CPS) methods stack images with \emph{time-multiplexing}, the target surface has to be kept static during the multiple shots. 
	With \emph{spectral-multiplexing}, multispectral photometric stereo~(MPS)~\cite{kontsevich1994} can recover surface normals from a one-shot multispectral image.
	As shown in \fref{fig:teaser}, a single input image for MPS encodes observations under varying lighting directions in different spectral bands, conveying the information about surface normals and spectral reflectances. With a multispectral camera and spectral light sources, MPS can reconstruct the 3D shapes of dynamic objects. 
	\footnotetext{Please check the supplementary video for the full sequence.}
	% step 2: Why it is important?
	Existing MPS methods~\cite{guo2021mps, chakrabarti2016single, ozawa2018single, anderson2011color} mostly focus on the Lambertian reflectance, which has limited ability to represent the surface appearance in real-world scenes.
% 	For non-Lambertian surfaces, the spectral reflectance is not only varying from the spectral band, but also the incident-outgoing lighting directions w.r.t the surface normal at the scene point. 
	For non-Lambertian surfaces, the spectral reflectance varies not only with the spectral band, but also the incident-outgoing lighting directions w.r.t. the surface normal at different scene points.
	There have been a few MPS methods working for non-Lambertian reflectance, which assumed a specific hardware setup~\cite{rahman2014color} with two-shot capturing or an input image containing only three spectral channels with fixed spectral bands~\cite{ju2018demultiplexing, ju2020dual, ju2020mps}. 
	Due to the restrictive capture setting and limited surface normal estimation accuracy of existing methods, non-Lambertian MPS remains an open and challenging problem. 
	
 Despite that modeling non-Lambertian reflectance is well-studied in the CPS context and great progress has been achieved with data-driven approaches (\eg,~\cite{chen2018ps, santo2017deep, yao2020gps, ikehata2018cnn}), non-Lambertian CPS methods cannot be directly applied to multispectral image observations, as the wavelength-dependent responses have not been considered.

% 	In this work, we propose a spectral reflectance decomposition~(SRD) model that enables the application of off-the-shell CPS methods on the MPS task to achieve accurate dynamic shape recovery on non-Lambertian surfaces.
	In this paper, we propose a spectral reflectance decomposition~(SRD) model for the task of MPS, which enables accurate recovery of dynamic non-Lambertian surfaces. With the proposed SRD model, MPS can be reformulated as a CPS problem, which allows us to borrow experiences from well-established CPS theory and practice.
	Specifically, we decompose the non-Lambertian spectral reflectance into two independent components: \emph{geometric} and \emph{spectral}. The geometric component only varies with the incident-outgoing lighting directions, which can be well-modeled by non-Lambertian CPS methods. The spectral component is the response w.r.t. the spectral wavelength. For surfaces covered by a uniform material, we can entangle it with the input light intensity as an \emph{equivalent light intensity}. Then the MPS problem can be formulated as a CPS one with unknown equivalent light intensities. 
	We, therefore, propose a neural network \textit{NeuralMPS} to first predict the equivalent light intensity, including the spectral component, and then embed existing CPS methods to recover the surface normal map considering the geometric component only. 
	
% 	We observe that the dynamic range of the equivalent light intensity can be very high (\eg, $[\num{1e-6}, 1]$), increasing the difficulty of intensity estimation using regression in the linear space or by lighting classification~\cite{chen2019self}.
% 	To deal with the high dynamic range (HDR) nature of the equivalent light intensity, we introduce an HDR regression loss based on the differentiable $\mu$-law function~\cite{kalantari2017deep} to achieve better results. 
% 	Extensive experiments on both synthetic and real datasets show that NeuralMPS outperforms existing non-Lambertian MPS methods. 
	To summarize, our contributions are as follows: 
	\begin{itemize}
	    \vspace{4.0pt}
        \item We introduce an SRD model and transform the MPS problem to the well-studied CPS with unknown equivalent light intensity.
	    \vspace{4.0pt}
		\item We propose a learning-based MPS network named NeuralMPS to obtain accurate surface normal prediction under non-Lambertian reflectance. %To the best of our knowledge, this is the first MPS method that 
	    \vspace{4.0pt}
        \item Extensive experiments on both synthetic and real datasets show that NeuralMPS outperforms existing non-Lambertian MPS methods. 
	    \vspace{4.0pt}
	\end{itemize}

	\begin{figure*}
		\resizebox{0.99\textwidth}{!}{
		\centering
		\scriptsize
		\begin{tabular}{@{}c|c|c|c}
			\Xhline{4\arrayrulewidth} 
			Lambertian, CPS & Non-Lambertian, CPS & Lambertian, MPS & Non-Lambertian, MPS
			\\
			$R(\V{n}, \V{l}, \lambda) = c$ & $R(\V{n}, \V{l}, \lambda) = R(\V{n}, \V{l})$ & $R(\V{n}, \V{l}, \lambda) = R(\lambda)$ & $R(\V{n}, \V{l}, \lambda) = R_s(\lambda)R_g(\V{n}, \V{l})$\\
			\Xhline{2\arrayrulewidth}
			{\includegraphics[ width=\imgsize\textwidth]{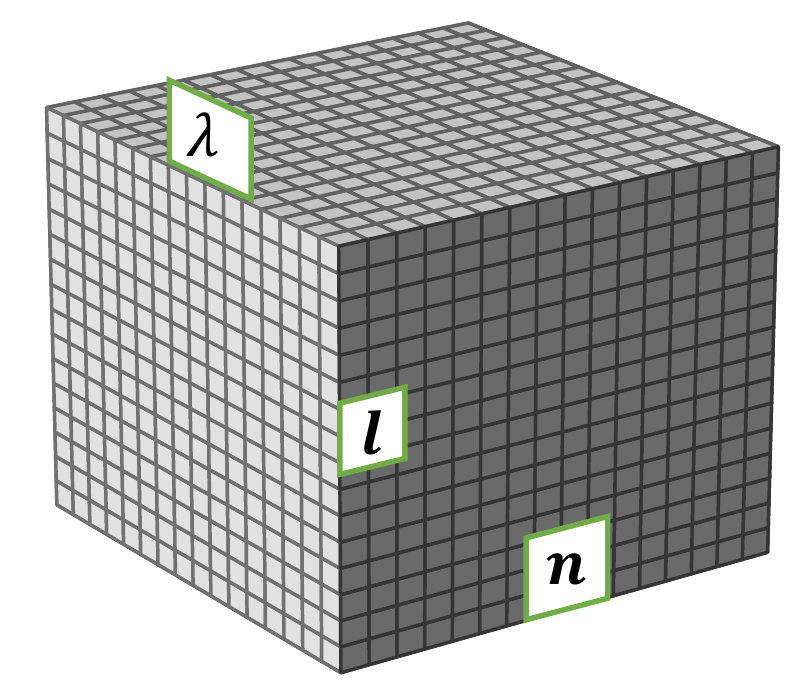}}
			&{\includegraphics[ width=\imgsize\textwidth]{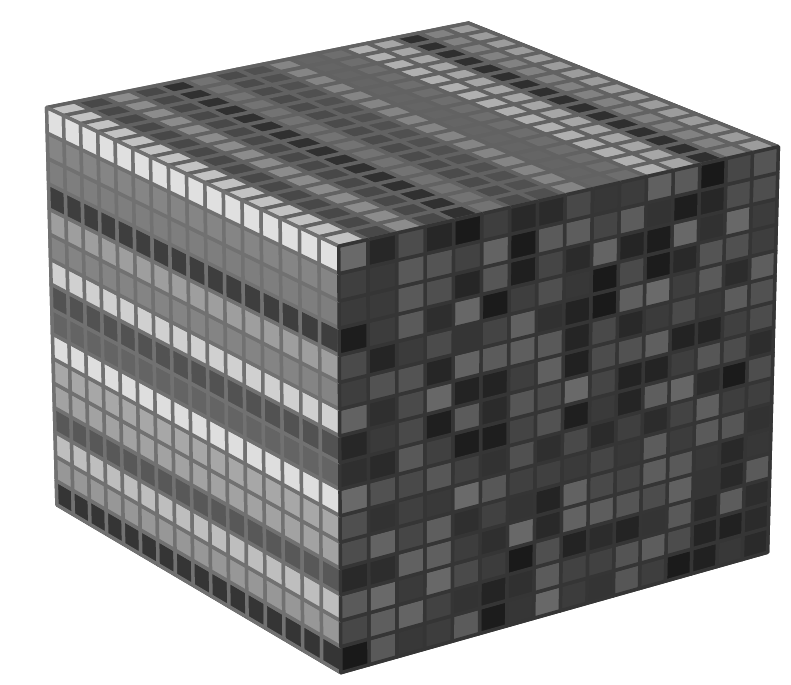}}
			&{\includegraphics[ width=\imgsize\textwidth]{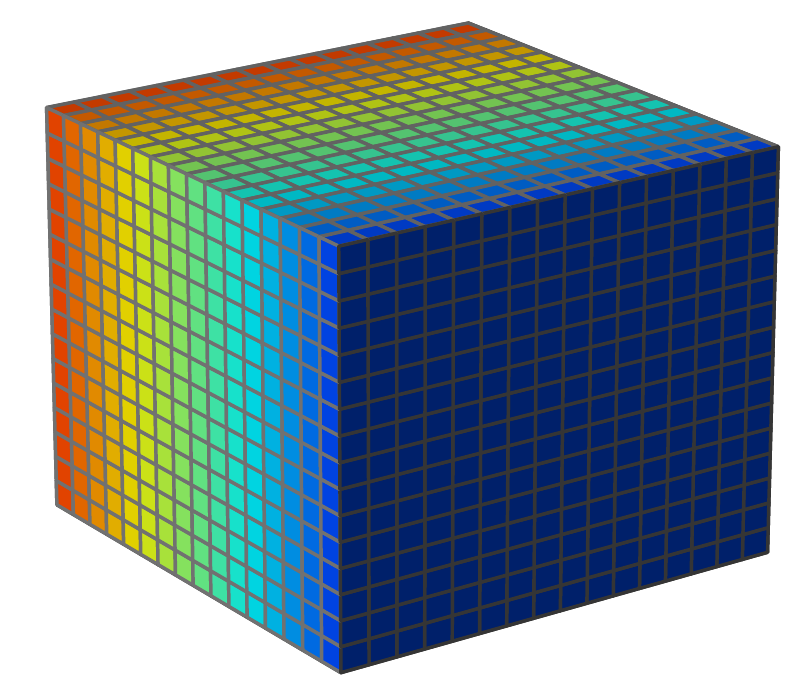}}
			&{\includegraphics[ width=\imgsize\textwidth]{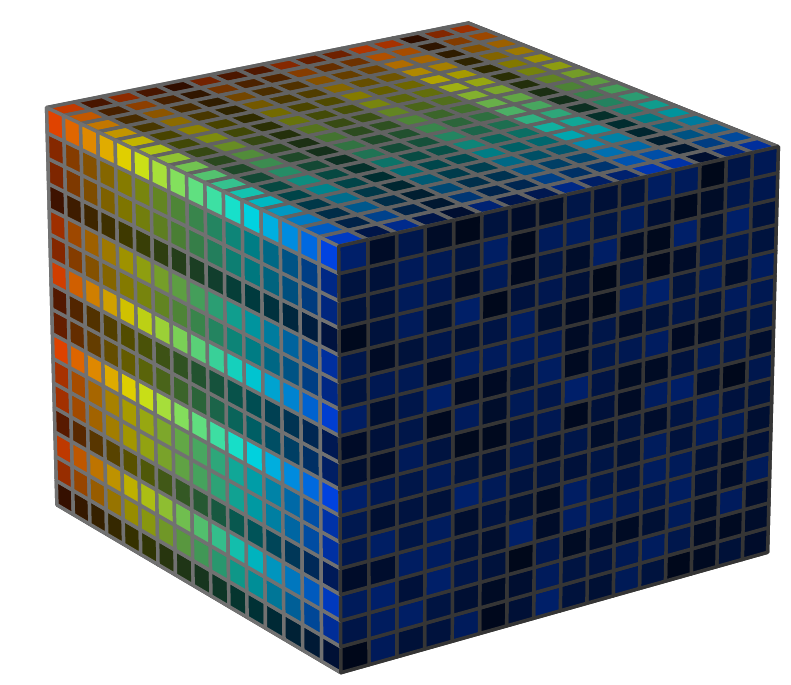}}\\
			\Xhline{2\arrayrulewidth}
			{\includegraphics[ width=\imgsize\textwidth]{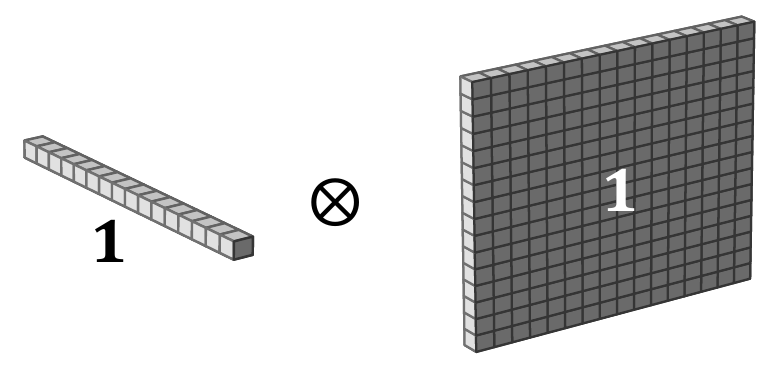}}
			&{\includegraphics[ width=\imgsize\textwidth]{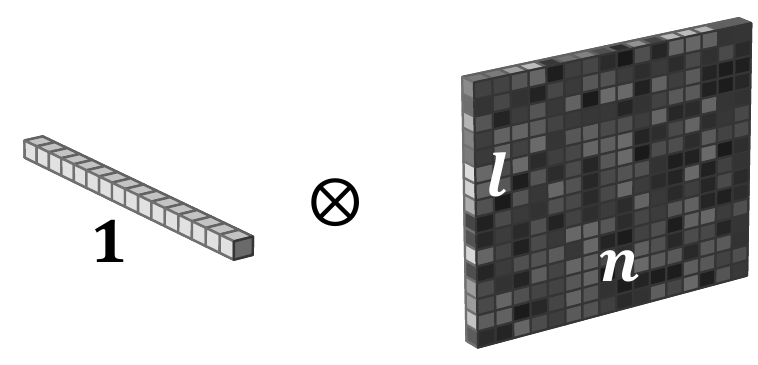}}
			&{\includegraphics[ width=\imgsize\textwidth]{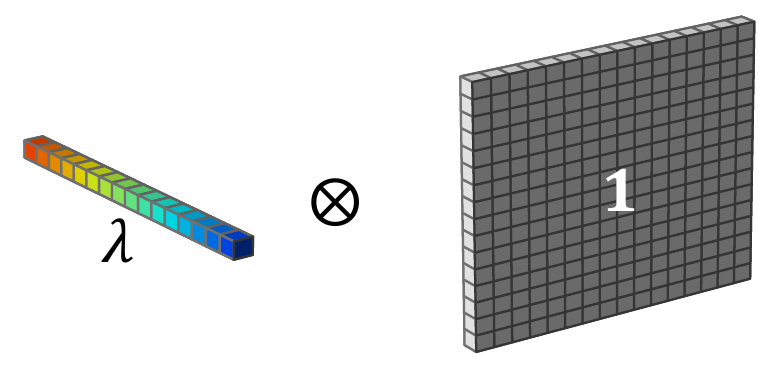}}
			&\includegraphics[ width=\imgsize\textwidth]{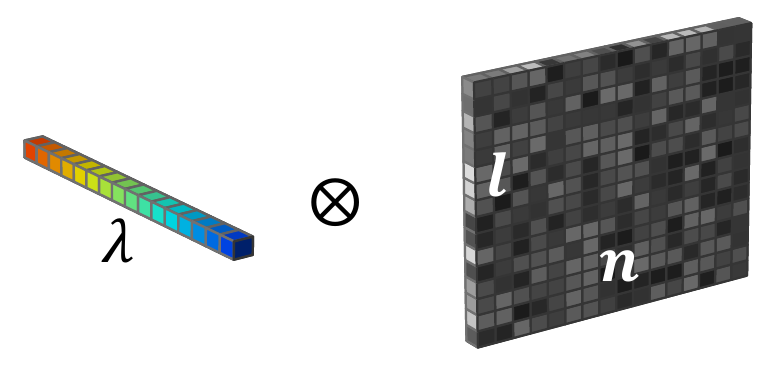}\\
			(a) & (b) & (c) & (d) \\
			\Xhline{4\arrayrulewidth}
		\end{tabular}
	}

	\caption{Visualization of the four categories of reflectance assumptions in existing photometric stereo methods, where $\otimes$ represents Kronecker product and $\V{1}$ denotes one vector/matrix whose elements are all equal to $1$. The reflectance decomposition in different photometric stereo tasks is shown in the third row. The last column illustrates the spectral reflectance decomposition model proposed in this paper. }
	\label{fig:sbrdf}
	\vspace{-1em}
	\end{figure*}
	\section{Related Works}
	
	The general reflectance for a surface can be modeled by the Bidirectional Reflectance Distribution Function~(BRDF)~\cite{szeliski2010computer}, which describes how much light at each wavelength arriving at an incident direction is emitted in a reflected direction. With a fixed view direction, the isotropic spectral BRDF is a function $R(\V{n}, \V{l}, \lambda)$ of the surface normal $\V{n}$, the lighting direction $\V{l}$ and the wavelength $\lambda$. Therefore, the BRDF response \wrt the three variables can be recorded as a cube, as shown in \fref{fig:sbrdf}.
    In this section, we review CPS and MPS approaches under Lambertian reflectance and non-Lambertian reflectance, respectively. 
    These four categories are based on different simplifications on the spectral BRDF cube.
	Among all the categories, we assume distant lights with calibrated directions.

	\paragraph{Lambertian, CPS}
	CPS under the Lambertian reflectance assumption is the most classic setup. 
    In such case, the spectral reflectance under varying lighting directions and wavelengths is a constant value, \ie, $R(\V{n}, \V{l}, \lambda) = c$ (see~\fref{fig:sbrdf}~(a)). Without loss of generality, we set $c = 1$.
	Given image measurements under more than $3$ non-coplanar lighting directions, classical photometric stereo works~\cite{woodham1980ps, silver1980determining} provided a closed-form solution to surface normal estimation.
	
	\paragraph{Non-Lambertian, CPS}
    The spectral reflectance for non-Lambertian CPS is varying with the incident-outgoing lighting directions (geometric component), but omitting the variation w.r.t the wavelength (spectral component). Therefore, the spectral BRDF $R(\V{n}, \V{l}, \lambda) = R(\V{n}, \V{l})$ and the spectral cube shown in \fref{fig:sbrdf}~(b) is a Kronecker product of an all-in-one spectral response vector and the geometric response matrix.
	Existing non-Lambertian CPS methods applied analytical~\cite{chen2019microfacet, shi2014bi} or non-parametric model~\cite{enomoto2020photometric, hui2016shape} to represent the geometric component of the spectral reflectance. Recently, neural-based photometric stereo methods~\cite{chen2018ps, ikehata2018cnn, yao2020gps} achieved high accuracy on surface normal estimation under non-Lambertian reflectance, where the geometric component is learned from data with diverse reflectances. Please refer to the survey~\cite{shi2019benchmark} for a comprehensive review of non-Lambertian CPS methods.
	
	\paragraph{Lambertian, MPS}
    In MPS, the spectral component has to be taken into account. Under Lambertian reflectance, the BRDF keeps constant w.r.t. geometric component, \ie, $R(\V{n}, \V{l}, \lambda) = R(\lambda)$. Therefore, the spectral cube shown in \fref{fig:sbrdf}~(c) is a Kronecker product of spectral response vector and an all-in-one matrix recording the variation on the geometric response. Different from Lambertian CPS, MPS is ill-posed even under Lambertian reflectance. Existing methods require additional shape priors~\cite{anderson2011color, anderson2011augmenting} or reflectance clustering~\cite{ozawa2018single, chakrabarti2016single} to solve the problem. Recently, Guo~\etal~\cite{guo2021mps} formulated the Lambertian MPS problem into a well-posed one and provided a unique solution for surface normals with image cue only.

	\paragraph{Non-Lambertian, MPS}
    Non-Lambertian MPS problem is the most challenging case among the four categories, as the spectral BRDF $R(\V{n}, \V{l}, \lambda)$ is related to both wavelength and incident-outgoing lighting directions. Therefore, the spectral BRDF is recorded as a general cube (see~\fref{fig:sbrdf}~(d)). Existing non-Lambertian MPS methods assumed analytical BRDF model and two-shot data capturing~\cite{rahman2014color}. However, the analytical BRDF model is limited to a small set of materials, and the two-shot data capturing eliminates the advantage of MPS over CPS on dynamic shape recovery. Ju~\etal~\cite{ju2020dual, ju2020mps} propose learning-based methods to solve non-Lambertian reflectance. However, the spectral band of the input multispectral images in the test phase is required to keep fixed in the training stage, which limits their application to real-world scenarios. To obtain accurate dynamic surface normal recovery under non-Lambertian reflectance, our method regards the spectral BRDF cube as a Kronecker product of the spectral response vector~(spectral component) and geometric response matrix~(geometric component). As we can see in the following sections, such a decomposition enables the application of off-the-shell non-Lambertian CPS methods on the non-Lambertian MPS task.

	\section{Spectral Reflectance Decomposition Model}
	\label{sec:srd}
	In the context of MPS, the material spectral reflectance varies with both the incident-outgoing lighting directions and spectral bands. Under a fixed view direction, it can be formulated as a function of surface normal $\V{n}$, lighting direction $\V{l}$, and wavelength $\lambda$. Our spectral decomposition model disentangles the non-Lambertian spectral reflectance into two independent components such that
	\begin{eqnarray}
		R(\V{n}, \V{l}, \lambda) = R_s(\lambda)R_g(\V{n}, \V{l}),
		\label{eq:spectral_decom}
	\end{eqnarray}
	where $R_s(\lambda)$ is the spectral component related only to the wavelength, and $R_g(\V{n}, \V{l})$ denotes the geometric component that coresponses to the variation w.r.t. surface normal and lighting direction. 
	For surfaces covered by uniform material, we assume all scene points share the same spectral component, but the geometric response can vary from point to point depending on their surface normal directions.
	Intuitively, this assumption means the chromaticity of the whole surface will not change by moving the light source positions. 
% 	Similar spectral reflectance decomposition has appeared in~\cite{anderson2011color, esteban2011overcoming}, but they focus on the Lambertian case, where $R_g(\V{n}, \V{l})$ keeps constant. 
	Similar assumptions about spectral reflectance have been used in~\cite{anderson2011color, esteban2011overcoming}, with a focus on the Lambertian case where $R_g(\V{n}, \V{l})$ keeps constant. 
	\begin{figure}
		\includegraphics[width = 1\linewidth]{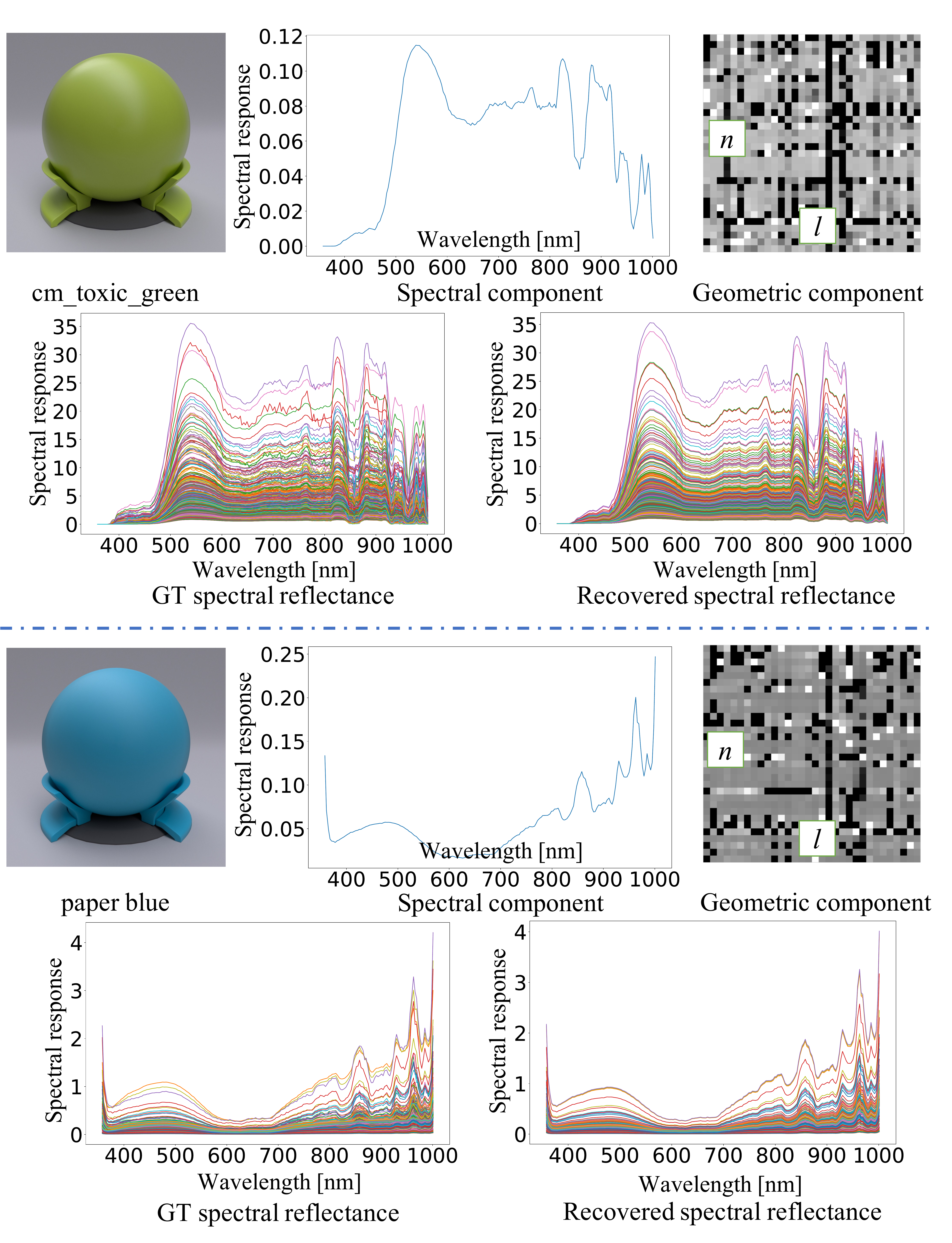}
        \caption{Illustration of spectral reflectance decomposition on two measured BRDFs ``cm\_toxic\_green'' and ``paper blue''~\cite{dupuy2018adaptive}. In each example, the top three figures show the material appearance, the spectral component and the geometry component map where each pixel corresponds to one pair of $(\V{l}, \V{n})$. The bottom row gives the ground-truth and recovered spectral reflectance.}
		\label{fig:spectral_decomp}
		\vspace{-1em}
	\end{figure}
	
	We illustrate and verify our SRD model by measured spectral BRDFs~\cite{dupuy2018adaptive}. As shown in \fref{fig:spectral_decomp}, we plot measured spectral reflectances of two materials: ``paper blue'' and ``cm\_toxic\_green'' in the $[360\text{nm}, 1000\text{nm}]$ wavelength range, where different color curves show the spectral reflectance at varying lighting direction and surface normal pairs: $(\V{l}, \V{n})$. 
	If the surface follows Lambertian reflectance~\cite{anderson2011color, esteban2011overcoming}, the spectral reflectance shown in the figure should contain one single curve, which is not flexible to represent the real-world spectral reflectances. 
	To verify our SRD model, we sample $195$ wavelengths and $200 \times 200$ pairs of $(\V{l}, \V{n})$ to build the spectral reflectance response matrix $\V{R} \in \mathbb{R}^{195, 200 \times 200}$. 
	To obtain decomposed reflectance component, we conduct SVD on $\V{R}$ such that
	\begin{eqnarray}
		\begin{aligned}
			\V{R} = \V{U} &\bm{\Sigma} \V{V}^\top, \\
			\V{r}_s = \V{U}_1, \quad &\V{r}_g = \V{V}_1,
		\end{aligned}
	\label{eq:svd_spectral}
	\end{eqnarray}
	where $\V{r}_s$ and $\V{r}_g$ are the decomposed spectral and geometric components in the vector form, which are obtained from the first column of \V{U} and \V{V}, respectively. 
	We observe that the largest singular value captures more than 75\% of the energy in $\bm{\Sigma}$, which reveals that we can approximate the spectral reflectance in high accuracy with the decomposed $\V{r}_s$ and $\V{r}_g$. The recovered spectral reflectance curves from the spectral and geometric components are shown in the figure, which is close to the originally measured one.
	
	\subsection{SRD-based Image Formation Model}
	Based on the proposed SRD model, we show that the MPS problem is equivalent to the CPS under unknown light intensity. This conclusion can be derived from the multispectral image formation model.
	
	Following the conventional practice, we assume an orthogonal multispectral camera with a linear radiometric response and $f$ spectral directional lights with calibrated lighting directions.
	By turning on all the spectral light sources, we can capture a single multispectral image with $f$ spectral bands recording measurements under varying lighting directions.
	Consider a non-Lambertian surface with the spectral reflectance modeled by a general isotropic BRDF, the image observation for each scene point on the surface under the $j$-th incoming lighting can be written as follows:
	\begin{eqnarray}
		\label{eq:img_f_1}
		m_j = e_j {\rm max}(\V{n}^\top\V{l}_j, 0) \int_{\lambda} R(\V{n}, \V{l}_j, \lambda) C_j(\lambda) E_j(\lambda)  \, d\lambda,
	\end{eqnarray}
	where $\V{n} \in S^2 \subset \mathbb{R}^3$ represents the unit surface normal vector, $\V{l}_j \in S^2 \subset \mathbb{R}^3$ and $e_j\in\mathbb{R}^+$ are lighting direction and radiance of the $j$-th light source, respectively, with its spectra defined by $E_j(\lambda): \mathbb{R}^+ \rightarrow \mathbb{R}^+$. The camera spectral sensitivity at the $j$-th spectral band is represented by $C_j(\lambda)$, and the attach shadow is modeled by $\rm max(\cdot)$.
	
	We assume the crosstalk between spectral bands is negligible~\cite{guo2021mps, chakrabarti2016single, ozawa2018single}, \ie, the observations under each spectral light can only be observed in its corresponding camera channel. Based on the negligible crosstalk and the spectral reflectance decomposition model shown in \eref{eq:spectral_decom}, the image observation can be re-written as
	\begin{eqnarray}
		\label{eq:img_f_2}
		m_j = e^\prime_j R_g(\V{n}, \V{l}_j) {\rm max}(\V{n}^\top\V{l}_j, 0),
	\end{eqnarray}
	and $e^\prime_j$ is defined by
	\begin{eqnarray}	
		\label{eq:img_f_3}
		e^\prime_j =   e_j \int_{\lambda \in \Omega_j} R_s(\lambda)  C(\lambda) E(\lambda) \, d\lambda,	
	\end{eqnarray}
	where $\Omega_j$ is the $j$-th spectral band of the corresponding spectral light and camera channel. 
	The camera spectral sensitivity, light spectra, material spectral component, and the light intensity are encoded in the $e^\prime_j$. If the material spectral component is uniform for the whole surface (\eg uniform material), $e^\prime_j$ keeps constant for all the surface points. Therefore, we name $e^\prime_j$ as the \textit{equivalent light intensity}.
	According to the multispectral image formation model shown in \eref{eq:img_f_2}, given the image measurement $m$ and its corresponding lighting direction $\V{l}$, the non-Lambertian MPS task can be reformulated as estimating the surface normal $\V{n}$ under unknown geometric reflectance $R_g(\V{n}, \V{l})$ and the equivalent light intensity $e^\prime_j $. In this way, {\it non-Lambertian MPS is equivalent to the non-Lambertian CPS problem with unknown light intensities}.

\begin{figure*}
	\includegraphics[width = 1\linewidth]{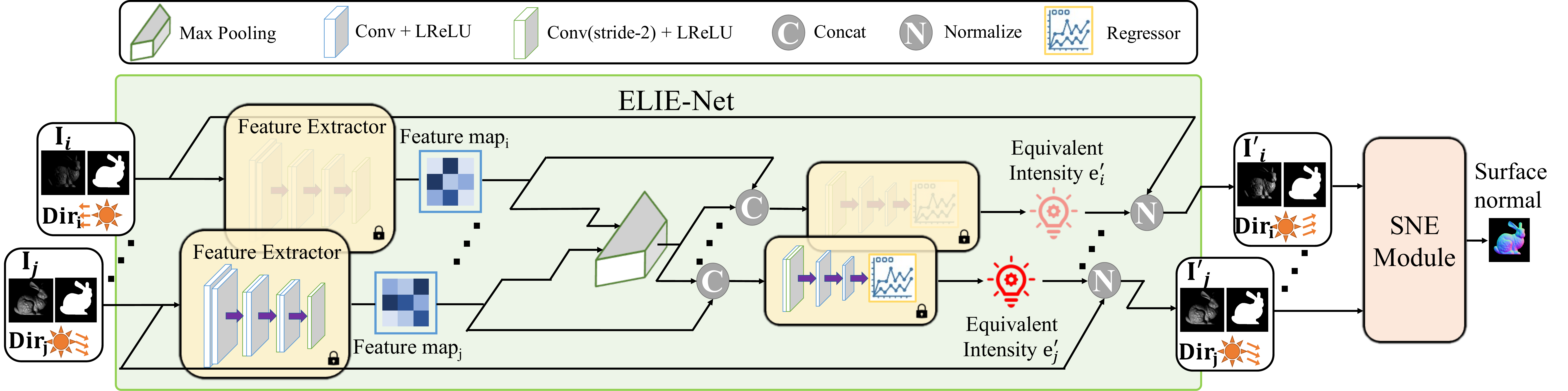}
    \caption{Network structure of our NeuralMPS. The proposed ELIE-Net is fed with image observations under varying calibrated lighting directions and the corresponding masks for estimating equivalent light intensity, from which the images are then normalized (denoted as $\mathbf{I'}$) and sent to the SNE Module for estimating surface normal. 
    For all image observations, the parameters are shared in the convolution layers with the ``lock" label. }
		\label{fig:network_structure}
		\vspace{-1em}
\end{figure*}

\section{Learning Non-Lambertian MPS}
In this section, we present our NeuralMPS for surface normal recovery under non-Lambertian reflectances. As shown in \fref{fig:network_structure}, there are two modules in the proposed method with functionalities indicated by their names: Equivalent Light Intensity Estimation Network (\lightNet) and Surface Normal Estimation (SNE) Module.

\begin{figure}\centering
	\includegraphics[width=1\linewidth]{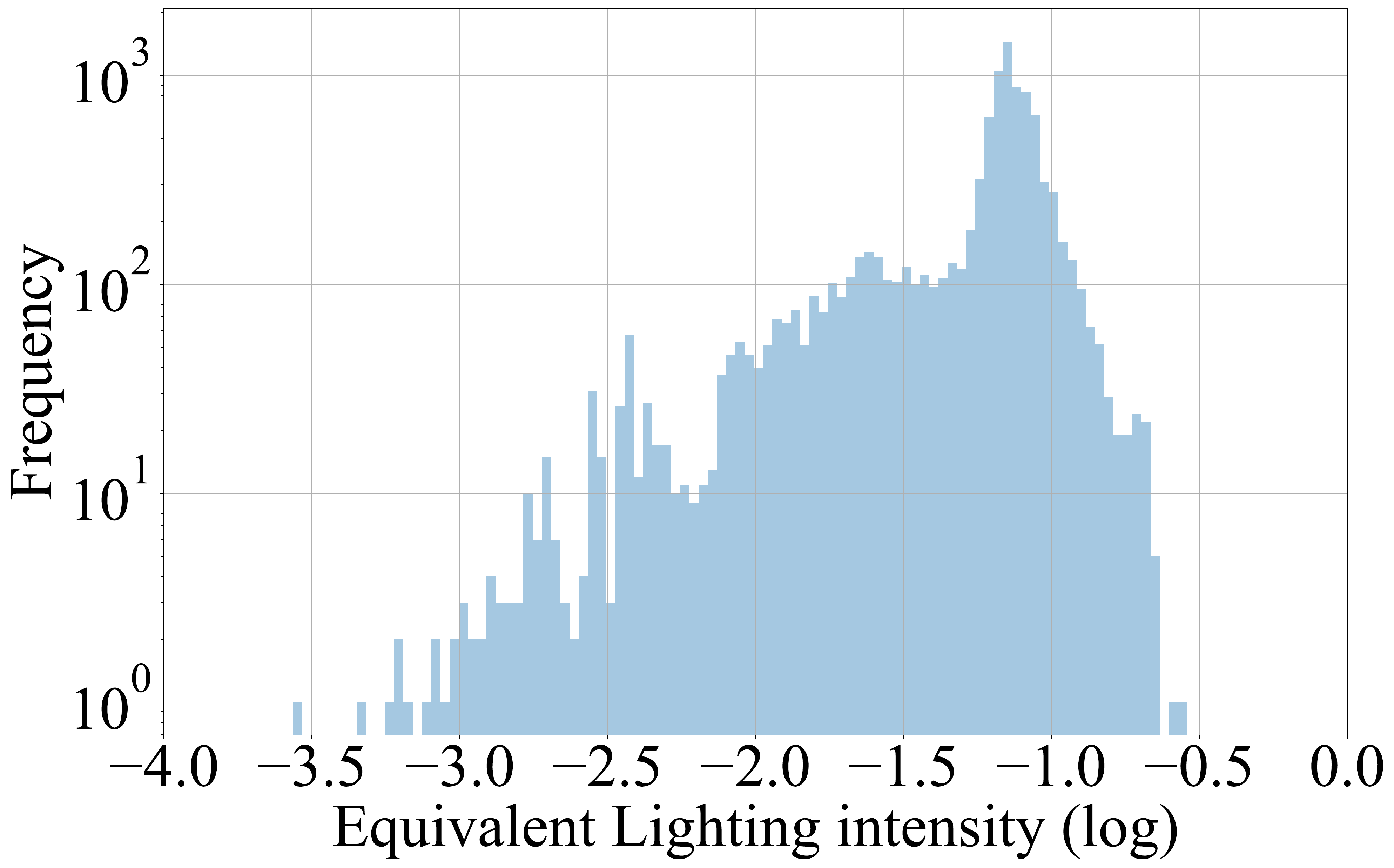}
	\caption{Histogram of equivalent light intensities has a broad range, which shows their property of high dynamic range.  
    \vspace{-1em}
	}
	\label{fig:light_hist}
\end{figure}
\paragraph{\lightNet}
As discussed in the previous section, the MPS problem is recast into a CPS with unknown equivalent light intensities. \lightNet aims at predicting the equivalent light intensities from the image observations and the calibrated lighting directions.
The overall framework is shown in \fref{fig:network_structure}.
Specifically, each image observation, as well as the corresponding calibrated lighting direction and object mask, are fed into a shared encoder separately to extract their local features first. 
The local features of different observations are then pooled into a global one by a max-pooling layer. By a shared decoder fed with concatenated local and global features, the equivalent light intensity for each observation can then be predicted.

The network architecture of \lightNet is similar to the LC-Net in~\cite{chen2019self}. 
However, the LC-Net in~\cite{chen2019self} is used to estimate both lighting direction and intensity in uncalibrated CPS, which is a different task from our MPS problem. 
Also, to ease the learning difficulty in their task, the regression-based problems of lighting direction and intensity are recast as a classification-based one, where the continuous lighting space is discretized into a discrete one with pre-defined ranges and classes.
However, we show that such discretization is too rough for the estimation of equivalent lighting intensity in our MPS task. Specifically, the equivalent light intensity in our method has a high dynamic range (HDR), as it encodes not only the irradiance of the light source but also the integral of camera spectral sensitivity, light spectra, and the material spectral component according to~\eref{eq:img_f_3}.
As shown in \fref{fig:light_hist}, we plot the histogram of equivalent light intensities from 51 measured materials, whose log intensity value range varies from $10^{-4}$ to $10^{-0.5}$, revealing the HDR property of equivalent light intensity.
Therefore, we treat the equivalent light intensity prediction from the image observations and calibrated lighting directions as a regression problem, which is different from the classification loss in LC-Net~\cite{chen2019self}. Given the ground-truth equivalent lighting intensity, we supervise the \lightNet by the $\ell_2$ regression loss:
\begin{eqnarray}
	\mathcal{L}_{ins} = \left\| \V{s} - \hat{\V{s}}\right\|_2^2,
\end{eqnarray}
where $\V{s}$ and $\hat{\V{s}}$ correspond to the GT and predicted equivalent lighting intensity.

\paragraph{\normalModule}
With the estimated equivalent lighting intensities, we can then normalize the input multispectral images at corresponding spectral bands with equal lighting intensities.
With these normalized image observations~($\V{I}'$ in \fref{fig:network_structure}) as well as the calibrated lighting directions as inputs, our \normalModule recovers the surface normal map with considering the geometric component only. 
As discussed in \sref{sec:srd}, any existing non-Lambertian CPS methods can be used as our \normalModule. 
In this paper, we discuss and test the state-of-the-art CPS method PS-FCN~\cite{chen2018ps} as our \normalModule.
For better collaboration between the proposed \lightNet and the \normalModule, the PS-FCN is re-trained on the proposed multispectral dataset ``PS-Spectral'', as we will show in the next. Following the training strategy of PS-FCN~\cite{chen2018ps}, the loss function is defined as the cosine similarity loss of the surface normal:

\begin{eqnarray}
	\label{eq:n_loss}
	\mathcal{L}_n = \frac{1}{p} \sum_{i=1}^p(1 - \V{n}_i^\top\hat{\V{n}}_i),
\end{eqnarray}
where $p$ is the number of valid pixels in the object mask, $\V{n}_i$ and $\hat{\V{n}}_i$ denote the ground-truth and estimated surface normal vectors at the $i$-th pixel position.

\paragraph{Difference from SDPS-Net}
As discussed above, our network structure is similar to SDPS-Net~\cite{chen2019self}. 
Here we elaborate on the differences between our work and SDPS-Net~\cite{chen2019self}. 
Firstly, SDPS~\cite{chen2019self} focuses on the non-Lambertian uncalibrated CPS problem, which has a completely different image formation from the non-Lambertian MPS problem that we focus on.
In this paper, we embed the distribution of spectral bands into equivalent light intensity mathematically to ease the difficulties in solving the non-Lambertian MPS problem.
Secondly, PS-FCN in~\cite{chen2018ps} is chosen as the \normalModule for illustration of the proposed method. In fact, any existing CPS methods can be utilized as \normalModule.
Thirdly, the network design of the prediction head of \lightNet is different from the LC-Net~\cite{chen2019self} due to their different purposes and inputs/outputs.

\paragraph{``PS-Spectral" Dataset}
\label{subsec:MPS Dataset}
% We built a synthetic dataset of shapes from PS-Blobby dataset as well as PS-Sculpture dataset for training usage. Each shape was rendered with 51 isotropic BRDFs from Spectral-BRDF dataset. To demonstrate the rendering process of each input data, let's denote the number of wavelength bands as $t$, of light directions as $f$, of pixels in the image as $p$. We first  render a matrix of shape $[f, p, t]$. Then we do SVD-decomposition of the very matrix to get the equivalent light intensity vector of shape $[t]$. Finally we choose random $t$ out of $f$ directions to form the final input image group of shape $[p, t]$, just like the setting of real data, each light direction correspond to one wavelength band.
For training and testing, we build a synthetic dataset with shapes from the Blobby dataset~\cite{johnson2011shape} as well as the Sculpture dataset~\cite{wiles2017silnet}, where each shape is rendered with 51 measured isotropic spectral BRDFs~\cite{dupuy2018adaptive}.

\begin{figure}
	\includegraphics[width=1\linewidth]{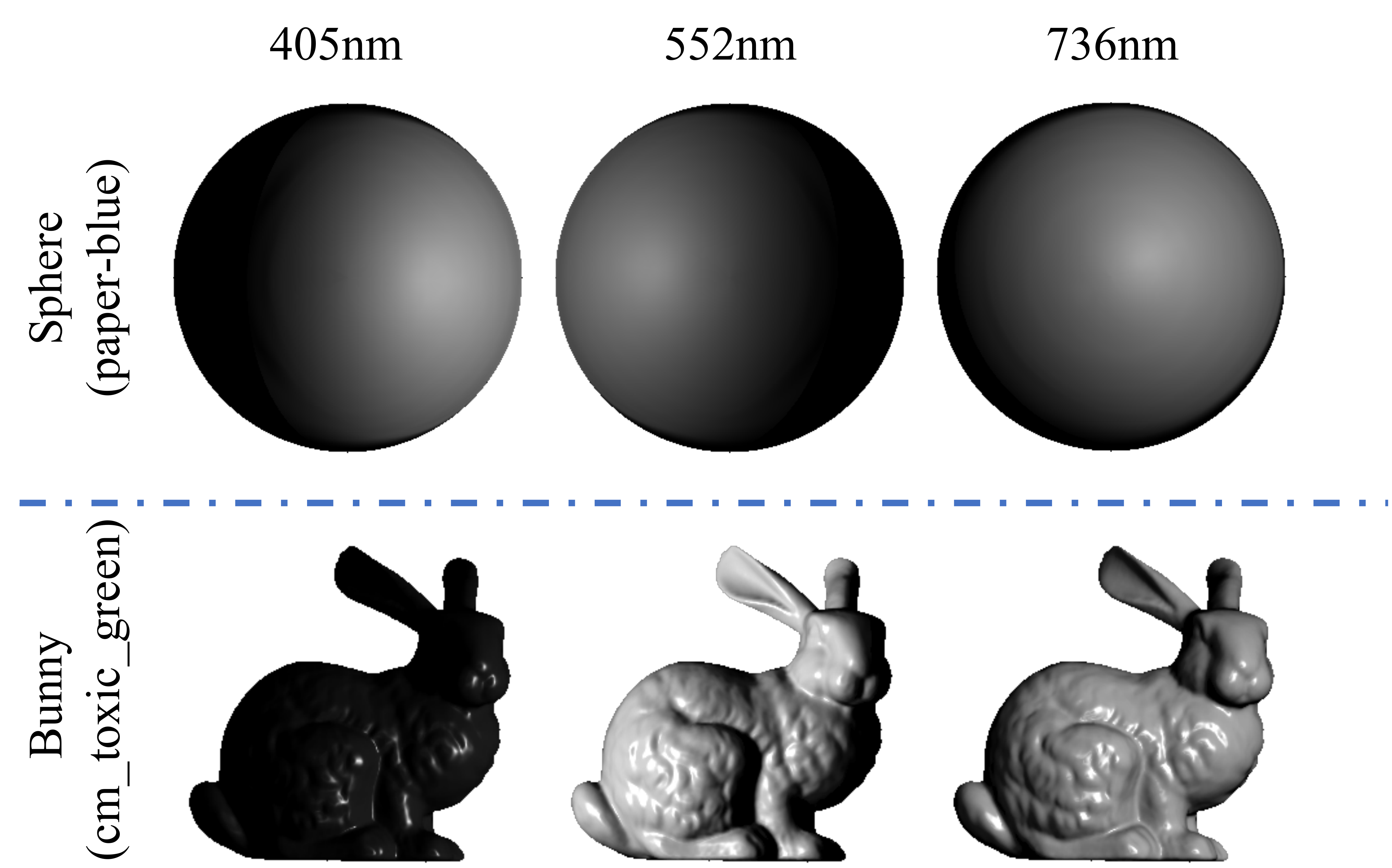}
	\caption{Synthetic data rendered under varying wavelength and lighting directions of {\sc Sphere} and {\sc Bunny} based on measured spectral BRDF dataset~\cite{dupuy2018adaptive}. }
	\label{fig:syndata}
	\vspace{-1em}
\end{figure}

For one object, the image observation tensor and the corresponding spectral reflectance tensor have the size of $[f, p, t]$, denoting the number of light directions, pixels in the image, and spectral bands, respectively. 
In MPS task, each lighting direction corresponds to one wavelength band, therefore we randomly select $t$ out of $f$ directions to produce one multispectral image data with the size of $[t, p, t]$. 
In this work, we select $f = 39$ lighting directions distributed uniformly, and $t=195$ spectral bands following the same wavelength range of~\cite{dupuy2018adaptive}. The image resolution in our rendering is set as $128 \times 128$.
To generate the ground-truth equivalent lighting intensity, we conduct the SVD decomposition on the spectral reflectance tensor $\V{R}$ with our SRD model. 
% \heng{Fill the number here}
We also render a test dataset including two shapes: {\sc Sphere} and {\sc Bunny}. The spectral image observations of the test dataset are shown in \fref{fig:syndata}. 

\paragraph{Implementation}
The two components \lightNet and \normalModule of our NeuralMPS are trained separately.
The ground truth equivalent light intensities are used as the supervised signal of \lightNet and the input of \normalModule. To increase the lighting invariance of the model, we simulated the randomly distributed lighting effects in the real world by a multiplication factor ranging in $(0.1,1)$ on the SVD light intensity during training.

\section{Experiments}
In this section, we evaluate our method on both synthetic and real captured data.
In \sref{subsec:srd_exp}, the accuracy of the proposed SRD model is verified on measured spectral BRDF database~\cite{dupuy2018adaptive} firstly.
Then, in \sref{subsec:ablation_exp} we compare our NeuralMPS with the state-of-the-art MPS method GO21~\cite{guo2021mps} on surface normal estimation on {\sc Bunny} and {\sc Sphere} test dataset. The comparison on surface normal recoveries with and without the \lightNet is also provided to evaluate effectiveness of our \lightNet module. 
Finally, in \sref{subsec:real_exp} we give the qualitative evaluation on real-captured data to verify our application on real scenarios.

\begin{figure*}\centering
	\includegraphics[width = 1\linewidth]{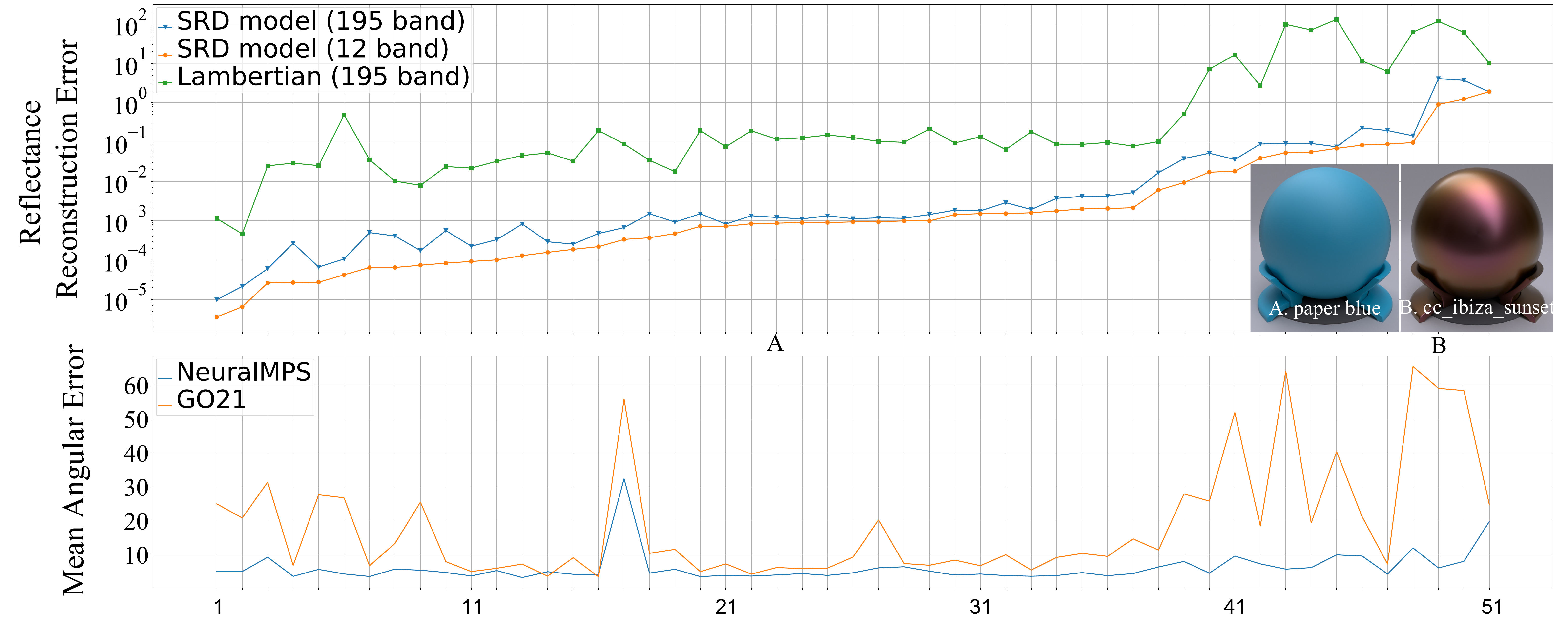}
	\caption{
		Evaluation of spectral reflectance decomposition for diverse materials~\cite{dupuy2018adaptive} (upper part).
		Evaluation of surface normal estimation for diverse materials~\cite{dupuy2018adaptive} (lower part).  The $X$-axis represents indices of the materials for both parts.}
	\label{fig:spec_decomp_eval}
	\vspace{-1em}
\end{figure*}
\subsection{Verification of the SRD Model}
\label{subsec:srd_exp}
% The SVD decomposition works well for most materials. 
Our SRD model decomposes the spectral reflectance into the geometric and spectral components via SVD decomposition. We verify the accuracy of the SRD model on diverse measured spectral BRDFs~\cite{dupuy2018adaptive}. Specifically, we first choose a sphere shape containing diverse surface normals and lighting directions with uniform distribution and then render image observations with $51$ spectral BRDFs~\cite{dupuy2018adaptive}. 
As mentioned in~\sref{subsec:MPS Dataset} and \eref{eq:svd_spectral}, the spectral reflectance can be concentrated as a matrix $\V{R}$ with shape $[f, p, t]$. With our SRD mode, $\V{R}$ is decomposed into geometric component $\V{R}_g$ of shape $[f, p]$ and spectral component $\V{r}_s$ of shape $[t]$ via SVD.
% ~\heng{Here it is inconsistent with the equivalent lighting intensity, but we need to use spectral component here as SRD is for spectral BRDF decomposition and not limited to MPS. We need to clarify this in the PS-Spectral dataset part.} 
In the case of spectral Lambertian assumption~\cite{guo2021mps}, the geometric component is an all-in-one matrix, $\V{R}_g = \V{1}$. Therefore, we find the spectral component $\V{r}_s$ in the Lambertian case that best fit $\V{R}_g$ and $\V{R}$ via least-square. 

We reconstruct the reflectance matrix \hbox{$\V{R}' = \V{R}_g \otimes \V{R}_s$} via tensor multiplication on the geometric and spectral component obtained by our SRD model and the Lambertian case. 
To measure reconstruction error $E_r$ over 51 measured materials, we use Frobenius norm between GT and reconstructed spectral reflectance, \ie, 
\begin{eqnarray}
	E_r = \frac{1}{51}\sum_i{\|\V{R}_i-\V{R}_i'\|_F}.
\end{eqnarray}
%distance matrix $\{ \Vert{M-M'}\Vert_F \}$ of all the 51 materials.
As shown in \fref{fig:spec_decomp_eval}~(top), the reconstructed spectral reflectance from our SRD model is more accurate than the Lambertian assumption used in GO21~\cite{guo2021mps}, revealing the effectiveness of our SRD model for general non-Lambertian spectral reflectance.
Besides, we found that the reconstruction errors with $t=12$ bands as well as $t=195$ bands have little difference. Therefore, our SRD model can be applied to real-world multispectral cameras with limited spectral bands. 

From \fref{fig:spec_decomp_eval} we observe that our SRD model works well for most materials. However, there are materials~(\eg, ``cc\_ibiza\_sunset'') showing varying spectral characteristics under different lighting directions, as shown on the top of the \fref{fig:spec_decomp_eval}. The spectral BRDFs for these materials violate our SRD model which assumes the spectral component is independent of the geometric component, like the monochromatic material ``paper\_blue''. In our NeuralMPS, we drop the last $9$ materials whose reconstruction errors are larger than $0.05$ during the training.

\subsection{Evaluation of NeuralMPS}

As shown in \fref{fig:spec_decomp_eval}~(bottom), we show the normal estimation result on the {\sc Sphere} test dataset.
Benefiting from our SRD model and designed network for non-Lambertian MPS, the surface normal estimation of our method is more accurate compared with the state-of-the-art MPS method GO21~\cite{guo2021mps} on diverse materials, revealing the strength of our method on general spectral reflectance.
% ~\heng{We drop $9$ materials but show the results. Good explanation for dropping materials is needed.}
%The result is consistent with our evaluation on the non-Lambertian reflectance shown in \fref{fig:spec_decomp_eval}.
%The results are shown in \fref{fig:spec_decomp_eval}. Here we compared our results with the state-of-the-art method GO21~\cite{guo2021mps}. % and CS16~\cite{chakrabarti2016single} as mentioned in \sref{subsec:baseline}. 
%% The comparison is conducted on the {\sc Sphere} and {\sc Bunny} dataset rendered with 51 materials. 
%As shown in \fref{fig:spec_decomp_eval}, for all the $51$ materials, the mean angular error of our surface normal estimates are smaller than the GO21~\cite{guo2021mps}, revealing the strength of our method on general spectral reflectance.

\paragraph{Effectiveness of \lightNet}
\label{subsec:ablation_exp}
As shown in \Tref{table:Ablation}, to demonstrate the effectiveness of \lightNet, we test our NeuralMPS with and without the module on synthetic data {\sc Sphere} and {\sc Bunny}. Specifically, ``w/o" or ``w/" in the first column refers to results generated by feeding \normalModule with all-one vector or equivalent light intensities estimated by ``\lightNet". Based on the mean angular error of the surface normal estimates over the test dataset,  the absence of equivalent light intensities leads to a performance drop in normal prediction. Therefore, the CPS method used in our SNE module~(``w/o \lightNet'') 
cannot be used for solving the MPS task. It is necessary to predict the equivalent lighting intensity including the spectral component of the spectral reflectance with our proposed \lightNet.

\begin{table}
	\caption{Evaluation of the effectiveness of \lightNet by the mean angular error of the test dataset.}
	\centering
	\begin{tabular}{cc|cc}
		\hline
		 \lightNet & \normalModule & {\sc Sphere} & {\sc Bunny}\\
		\hline
		 w/o  & PS-FCN~\cite{chen2018ps} & 7.05 & 13.12 \\
		 w/ & PS-FCN~\cite{chen2018ps} & 6.27 & 12.89 \\   % LDR: 6.27 12.89 
	\hline
	\end{tabular}
%	\boxin{This table could be more compact or even a single column one to save space? Sphere mean and others occupy too much space. For all tables, put captions on top of it.}
	\label{table:Ablation}
\end{table}

% Experiment 2 and 3 show the HDR loss can lead to better performance for lighting estimation, which is also evident in \fref{fig:light_hist}. 
% It means that training a \lightNet with HDR loss to predict the equivalent light intensities is more suitable for our MPS task.
% Compare experiments with IDs 1 and 3, we can see the absence of equivalent light intensities will lead to worse performance in normal prediction. Experiments 2 and 3 shows the HDR loss can lead to better performance in stage1, thus the MAE of . This is caused by the high dynamic range of SVD decomposition results, we illustrate this in \fref{fig:light_hist}. From comparisons among these three baseline models, we come to the conclusion that training a \lightNet with HDR loss to predict the equivalent light intensities is most suitable for our MPS task.

% \paragraph{Effectiveness of \normalModule}
% \label{subsec:NENet_eval}
% To validate the effectiveness of the normalization \jx{what dose `normalization' mean?} of the estimated equivalent light intensities for the reformulation of MPS to CPS, we compared the proposed methods with a state-of-the-art non-Lambertian CPS method \cite{Enomoto_2020_CVPR}, which is non-learning.
% As shown in the table, the proposed method achieves superior performance. Comparing also with Experiment 3 we can conclude that the learning-based method works better in normal map prediction as well as non-Lambertian CPS problem.

\begin{figure*}[]
	\vspace{3pt}
	\includegraphics[width = 0.97\linewidth]{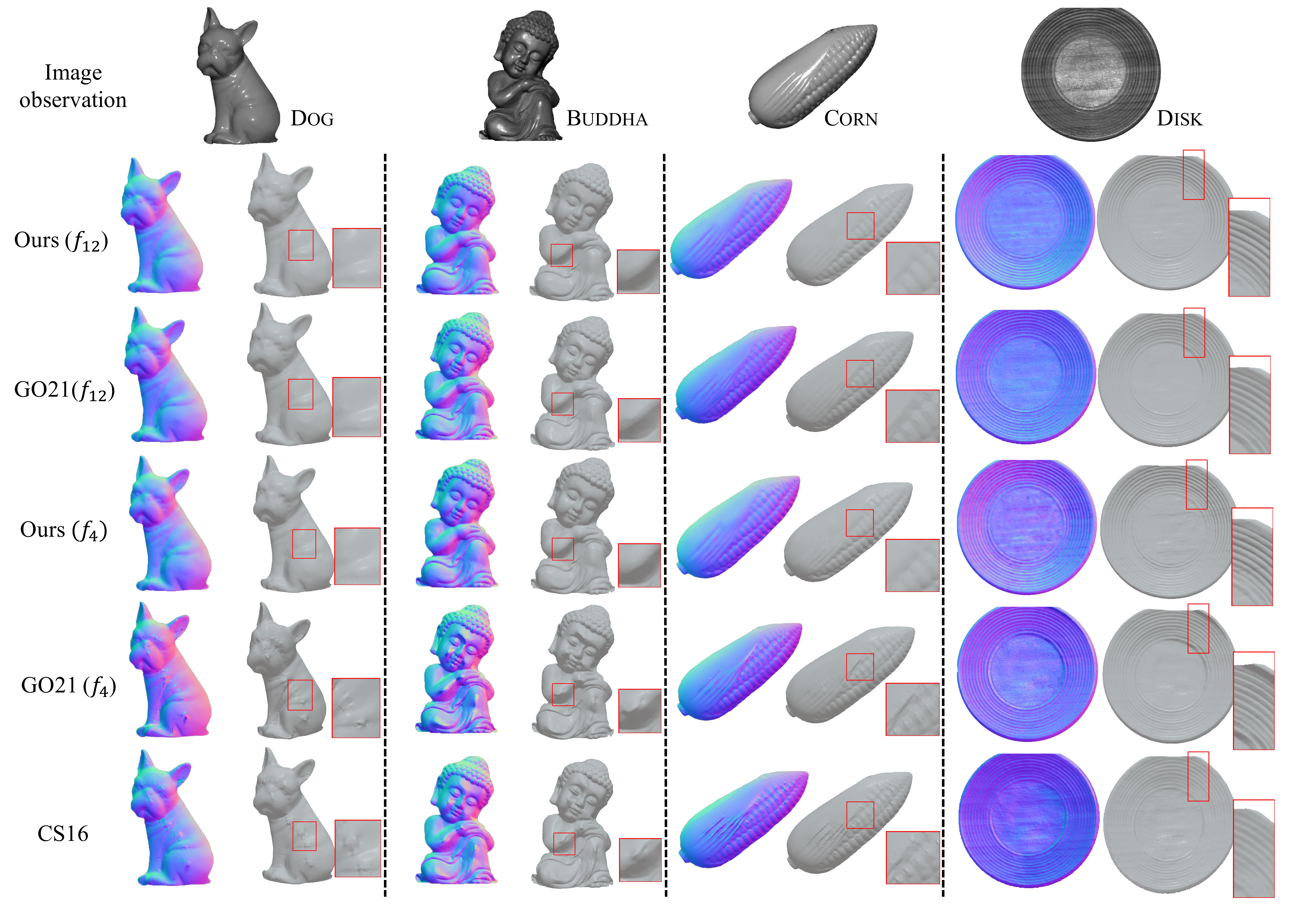}
	\vspace{2pt}
	\caption{Qualitative comparison with existing MPS methods on real data. Shape distortions caused by non-Lambertian specular highlights are emphasized in the closed-up views. 
% 	As shown in the figure, the results of GO21~$(f_4)$ and  contains strong artifacts caused by spectral reflectances. While our resul
	}
	\label{fig:Realdata_result}
	\vspace{-1em}
\end{figure*}

\subsection{Experiments on Real Data}
\label{subsec:real_exp}
As shown in \fref{fig:Realdata_result}, We test our method on real multispectral images of non-Lambertian surfaces captured by GO21~\cite{guo2021mps}. 
Taking the spectral image observations and calibrated lighting directions as input, we compare our NeuralMPS with existing MPS methods CS16~\cite{chakrabarti2016single} and GO21~\cite{guo2021mps} on surface normal recovery.
As CS16~\cite{chakrabarti2016single} takes only three-channel RGB images as input, we select the image observations with the wavelength 450, 550, and 650 [nm] to mimic the RGB input.
For GO21~\cite{guo2021mps} and our method, we test the methods on multispectral images with 12 and 4 spectral bands~(labeled as $f_4$ and $f_{12}$) to evaluate the application on real-world multispectral cameras with varying spectral channels.
Since the ground-truth surface normal and shape are not available for the real captured data, we conduct a qualitative evaluation by comparing the integrated shape from the surface normal map following~\cite{xie2014surface}.

As shown in \fref{fig:Realdata_result}, we show the estimated surface normals from existing methods and ours.
As CS16~\cite{chakrabarti2016single} mainly focuses on Lambertian reflectance, the recovered shape is distorted due to the specular highlights, as highlighted in the closed-up views.
GO21~\cite{guo2021mps} is adapted to arbitrarily many spectral bands as the input and removes the specular highlights and shadows as outliers with the position thresholding strategy~\cite{shi2019benchmark}. Therefore, the shape recoveries from GO21~$(f_{12})$ are reasonable with $12$ spectral bands as input. However, the method fails with limited spectral input~(GO21~$(f_4)$), as the position thresholding strategy prefers more data for the outlier removal.
Compared with existing methods, our NeuralMPS achieves reasonable surface normal estimation results for all four objects. The recovered shapes have no distortions caused by specular highlights. 
Since our method learns the non-Lambertian spectral reflectance from the measured spectral BRDF dataset, rather than treating it as outliers, the recovered surface normal under limited spectral bands remains in good quality, as shown in Ours~$(f_4)$. 

\section{Conclusion}
In this paper, we propose a multispectral photometric stereo method NeuralMPS for surface normal recovery under non-Lambertian spectral reflectance. The spectral reflectance decomposition model is the key for our method, which represents spectral reflectance as a composition of geometry components and spectral components. In this way, the non-Lambertian MPS problem can be recast into the well-studied non-Lambertian CPS problem with unknown equivalent light intensity in which the spectral component is embedded. As a result, we design the \lightNet and the \normalModule to estimate the equivalent light intensity and further recover the surface normal in MPS. 

\vspace{11pt}
\paragraph{Limitation}
We assume the surface is covered by uniform material so that the spectral reflectance can be decomposed into a global spectral component and geometric component. Intuitively, our NeuralMPS works on monochromatic targets as shown in \fref{fig:Realdata_result}. It is desirable to consider more general spatially-varying non-Lambertian spectral reflectance. As real-world surfaces are mostly covered by limited materials, one possible solution is adopting material classification and segmenting the surface into regions with uniform materials. Our NeuralMPS can be then applied for surface normal recovery for each sub-region.

%\begin{table}
%	\input{table/method_compare}
%	\caption{Category of calibrated photometric stereo methods \wrt reflectance type and capture setup.}
%	\label{tab:method_comp}
%\end{table}

\clearpage

{\small
	\bibliographystyle{ieee_fullname}
	\bibliography{heng}
}
%--------------------------------------------------------------------------------------------------

\end{document}